\documentclass[nonatbib,preprint]{elsarticle}

\usepackage[utf8]{inputenc} 
\usepackage[T1]{fontenc}    
\usepackage[hidelinks]{hyperref}       
\usepackage{url}            
\usepackage{booktabs}       
\usepackage{amsfonts}       
\usepackage{amssymb}        
\usepackage{amsmath}        
\usepackage{nicefrac}       
\usepackage{microtype}      
\usepackage{graphicx}
\usepackage{cleveref}
\usepackage{subcaption}
\usepackage{tabularx}
\usepackage{xcolor}
\hypersetup{
    colorlinks,
    linkcolor={red!50!black},
    citecolor={blue!50!black},
    urlcolor={blue!80!black}
}
\usepackage{tikz}
\usepackage{pgf}
\usetikzlibrary{matrix,positioning,calc,intersections,decorations,chains,decorations.pathreplacing,tikzmark}
\usepackage{pgfplots}
\usepackage{pgfpages}
\usepackage{pgfkeys}

\crefformat{equation}{(#2#1#3)}
\crefrangeformat{equation}{(#2#1#3) -- (#5#4#6)}
\crefmultiformat{equation}{(#2#1#3)}{ and~(#2#1#3)}{, (#2#1#3)}{ and~(#2#1#3)}

	\title{Continuous Learning and Adaptation with Membrane Potential and Activation Threshold Homeostasis}

\usepackage[australian]{babel}
\usepackage[style=australian]{csquotes}

\usepackage[%
backend=biber,
citestyle=numeric-comp,
bibstyle=numeric,
useprefix=false,
minbibnames=2,%
maxbibnames=3,%
maxcitenames=3,%
uniquename=true,
firstinits=true,
isbn=false,
doi=false,
sorting=nyt,
sortcites=true,
url=false,%
natbib=true,%
urldate=comp,
arxiv=abs,
alldates=year
]{biblatex}

\renewbibmacro{in:}{}

\AtEveryBibitem{%
	\ifentrytype{inproceedings}{
		\clearlist{publisher}
		\clearfield{volume}
		\clearfield{pages}
	}{}
	\clearlist{address}
	\clearname{editor}
	\clearfield{isbn}
	\clearfield{issn}
	\clearfield{doi}
	\clearlist{location}
	\clearfield{month}
	\clearfield{day}
	\clearfield{series}
	\clearfield{primaryclass}
	\clearfield{eprintclass}
	\clearfield{url}
	\clearfield{urlyear}
}

\newcommand{\norm}[1]{\left\lVert#1\right\rVert}

\pgfplotsset{compat=1.8}

\pgfplotsset{axis line style={black}}

\usepackage{tikz}
\usetikzlibrary{matrix,positioning,calc,intersections,decorations,chains,decorations.pathreplacing}

\author[1]{Alexander Hadjiivanov\corref{cor1}%
\fnref{fn1}}
\ead{alex@akin.com}

\affiliation[1]{organization={Core AI Research Division, Akin LABS},
city={Sydney},
postcode={NSW 2000},
country={Australia}}
\begin{document}

\begin{abstract}
	Most classical (non-spiking) neural network models disregard internal neuron dynamics and treat neurons as simple input integrators. However, biological neurons have an internal state governed by complex dynamics that plays a crucial role in learning, adaptation and the overall network activity and behaviour. This paper presents the Membrane Potential and Activation Threshold Homeostasis (MPATH) neuron model, which combines several biologically inspired mechanisms to efficiently simulate internal neuron dynamics with a single parameter analogous to the membrane time constant in biological neurons. The model allows neurons to maintain a form of dynamic equilibrium by automatically regulating their activity when presented with fluctuating input. One consequence of the MPATH model is that it imbues neurons with a sense of time without recurrent connections, paving the way for modelling processes that depend on temporal aspects of neuron activity. Experiments demonstrate the model's ability to adapt to and continually learn from its input.
\end{abstract}

\maketitle

\section{Background}\label{mpath.background}

The vast majority of neural network (NN) models (including popular deep learning models used extensively for applications ranging from self-driving cars to language understanding) are idealised and simplified models of a biological brain. This simplification is necessary because it is currently impossible to run a full-scale realistic simulation of the dynamics of a real brain due to the complex interactions occurring within and between neurons at every moment.

The human brain contains close to $ 10^{11} $ neurons, each of which communicates with an average of $ 10^{4} $ other neurons, putting the total number of neuron-neuron connections at close to $ 10^{15}$ \parencite{Herculano-Houzel--2009--HumanBrainNumbers,LaughlinSejnowski--2003--CommunicationNeuronalNetworks}. Despite this intimidating complexity, the biological brain makes use of a number of flexible mechanisms to ensure that it maintains a stable overall activity. Some of the more well-known mechanisms include long-term potentiation (LTP) and depression (LTD) (collectively known as Hebbian learning or spike-time-dependent plasticity (STDP)), which operate by modifying the synaptic efficacy depending on the activity of both presynaptic and postsynaptic neurons \parencite{Hebb--1949--OrganizationBehaviorNeuropsychological,BlissCollingridge--1993--SynapticModelMemory,Feldman--2012--SpikeTimingDependencePlasticity}. While LTP and LTD are to a certain extent \textit{extrinsic} in the sense that they affect presynaptic as well as postsynaptic neurons, the brain also employs a range of \textit{intrinsic} mechanisms that operate at the level of individual neurons.

For instance, neurons are capable of regulating their own spiking rate by employing homeostatic balancing mechanisms that regulate the neuron's activity depending on whether it is receiving too little or too much input. One such mechanism uses the relative number of $ Na $ and $ K $ channels to regulate the intrinsic excitability of the neuron \parencite{Turrigiano--2011--TooManyCooks,Davis--2006--HomeostaticControlNeural}. Another mechanism involves the modification of synaptic efficacy from the postsynaptic side by dynamically modulating the neuron's sensitivity to neurotransmitters released by presynaptic neurons \parencite{OBrienKambojEhlersEtAl--1998--ActivityDependentModulationSynaptic,LissinGompertsCarrollEtAl--1998--ActivityDifferentiallyRegulates}. Yet another mechanism, which activates when a new synapse is formed, scales the signals received through existing synapses in such a way that the total excitation after the addition of the new synapse remains stable \parencite{Turrigiano--2008--SelfTuningNeuronSynaptic,HartmannMinerTriesch--2016--PreciseSynapticEfficacy}. Although the synaptic scaling mechanism resembles that of LTD, is not the same since LTD affects a single synaptic weight, whereas synaptic scaling affects all of the neuron's inputs.

A number of homeostatic processes are also observed in the retina, which is composed of several layers of neurons involved in processing visual information. The photosensitive layer contains cells which can be broadly classified into cones (responsible for colour vision) and rods (which do not differentiate colours but are much more sensitive to light). Cones have a very large dynamic range and are primarily active in photopic vision (active during the day, when there is enough light). Cones are also extremely sensitive to small changes in contrast, both in terms of response time and contrast range. In comparison, rods are responsible for scotopic vision (activated in dark environments) and can detect very low levels of illumination (as low as individual quanta of light \parencite{MeisterBerry--1999--NeuralCodeRetina,BurnsLamb--2004--VisualTransductionRod,}), with the trade-off that they adapt very slowly and become saturated by large sudden changes in light intensity \parencite{BurnsLamb--2004--VisualTransductionRod}.

Signals produced by rods and cones are fed into two types of bipolar cells (BCs) (ON- and OFF-type), which have a characteristic center/surround structure. ON-type (OFF-type) BCs have a light center/dark surround (dark center/light surround) structure and respond strongly to positive (negative) contrast. The ON/OFF differentiation is maintained in retinal ganglion cells (RGCs), which receive input from one of the two types of BCs. The receptive fields of both ON and OFF RGCs independently cover almost the entire retina \parencite{RatliffBorghuisKaoEtAl--2010--RetinaStructuredProcess}. The retina uses adaptation mechanisms in BCs and RGCs to adapt to both spatial and temporal variations in illumination \parencite{BaccusMeister--2002--FastSlowContrast,FreemanGranaPassaglia--2010--RetinalGanglionCell,Rieke--2001--TemporalContrastAdaptation}. Both BCs and RGCs homeostatically regulate the intensity of their output depending on the magnitude of the stimulus, allowing the retina to adapt to changes in luminance that spans $ 9 $ to $ 12 $ orders of magnitude over a 24-hour period \parencite{RiderHenningStockman--2019--LightAdaptationControls,RiekeRudd--2009--ChallengesNaturalImages,PearsonKerschensteiner--2015--AmbientIlluminationSwitches}.

Importantly, there is overwhelming evidence that RGCs adapt their activity to both the mean and the variance of the input \parencite{Demb--2008--FunctionalCircuitryVisual,FreemanGranaPassaglia--2010--RetinalGanglionCell,MeisterBerry--1999--NeuralCodeRetina,BarlowLevick--1969--ChangesMaintainedDischarge,BaccusMeister--2002--FastSlowContrast,KimRieke--2001--TemporalContrastAdaptation,RiekeRudd--2009--ChallengesNaturalImages,RiderHenningStockman--2019--LightAdaptationControls}. In addition to temporal adaptation, both ON and OFF BCs also maintain a fast ``push-pull'' mechanism mechanism that balances the activation of the center relative to the surround for the purpose of locally enhancing the contrast within the receptive field \parencite{MeisterBerry--1999--NeuralCodeRetina,Enroth-CugellRobson--1966--ContrastSensitivityRetinal,VanWykWassleTaylor--2009--ReceptiveFieldProperties}. There is even some evidence that RGCs also employ a gain control mechanism to change the effective \textit{size} of the receptive field based on the illumination \parencite{BarlowFitzhughKuffler--1957--ChangeOrganizationReceptive}.

Another adaptation mechanism involves the neuron's spiking threshold. Biological neurons communicate with spikes, which are generated if the neuron's membrane is sufficiently depolarised by incoming spikes from presynaptic neurons. A spike is generated when the membrane potential crosses a certain threshold, which was initially assumed to be fixed at approximately $ -50 mV $ (compared to the resting potential of about $ -70 mV $). However, it has been shown that the spiking threshold is in fact also a variable quantity that is directly proportional to the neuron's mean membrane potential and inversely proportional to the speed of membrane depolarisation caused by presynaptic activity \parencite{AzouzGray--2003--AdaptiveCoincidenceDetection,AzouzGray--2000--DynamicSpikeThreshold,AzouzGray--1999--CellularMechanismsContributing,FontainePenaBrette--2014--SpikeThresholdAdaptationPredicted,HenzeBuzsaki--2001--ActionPotentialThreshold,PlatkiewiczBrette--2011--ImpactFastSodium}. This is interpreted as evidence that individual neurons act as \textit{coincidence detectors} that respond more readily to input spikes arriving in quick succession (with a small temporal delay) than to ones arriving further apart. Interestingly, changes in the spiking threshold seem to happen about an order of magnitude \textit{faster} than changes in the membrane potential \parencite{FontainePenaBrette--2014--SpikeThresholdAdaptationPredicted}. This mechanism effectively implements a high-pass filter that makes the neuron less sensitive to low-frequency fluctuations in the membrane potential while retaining its sensitivity to high-frequency ones. Certain models suggest that an adaptive threshold improves the robustness of decoding information encoded in the neuron's response \parencite{HuangResnikCelikelEtAl--2016--AdaptiveSpikeThreshold}, as well as that it makes the neuron less sensitive to the mean of the input and more sensitive to its variance\parencite{PlatkiewiczBrette--2011--ImpactFastSodium}.

In summary, all of the abovementioned mechanisms allow the neuron `tip the scales' in favour of excitation or inhibition, to `turn the volume' up and down on the received input, to `dampen' all input channels or to `shift the goalpost' for triggering an action potential, giving each neuron a wide (and quite possibly redundant) range of tools to regulate its own activity. Currently, classical NN models completely disregard the fact that the brain consists of many different types of neurons with their own specialised function and complex dynamics. The overwhelming majority of connectionist research places a strong focus on weight optimisation, which is certainly important but does not paint the full picture of how the biological brain maintains a stable activity, adapts to gradual or sustained changes in environmental stimuli and reacts to unpredictable deviations thereof. While spiking NN models provide a more accurate representation of their biological counterparts (such as the Hodgkin-Huxley \parencite{HodgkinHuxley--1952--QuantitativeDescriptionMembrane} or Izhikevich \parencite{Izhikevich--2003--SimpleModelSpiking,Izhikevich--2004--WhichModelUse}), they attempt to comply with the biological plausibility of many parameters and input modalities influencing the particular dynamics of neurons. As a side effect, many such models are computationally heavy (see, however, \parencite{Izhikevich--2004--WhichModelUse} for a comparison of computationally efficient simulation models). Therefore, what we are interested in is bridging this gap between spiking and classical models with an efficient, accurate and flexible model of neuronal equilibrium with a minimal dependence on additional parameters. A desirable property of such a model is \textit{genericity}, which means not only that it should it be able to handle any type of input, but also that it should handle input values which are not confined to biologically meaningful ranges.

This paper contributes a biologically inspired model of the above adaptation mechanisms, with the addition of only \textit{one} parameter (analogous to the membrane time constant in biological neurons) compared to classical NN models. The model is presented in Section \ref{mpath.model}, followed by simulation results in Section \ref{mpath.experiments} that demonstrate the model's ability to maintain homeostatic equilibrium in terms of neural activity. A discussion of potential limitations and possible directions of future research are given in Section \ref{mpath.discussion}.

\section{Membrane Potential and Activation Threshold Homeostasis model}\label{mpath.model}
As outlined above, neuronal activity can be kept at equilibrium by means of several mechanisms, which must operate in synchrony in order to not introduce additional chaos into the already highly fluctuating input faced by a neuron. The membrane potential and activation threshold homeostasis (MPATH) model consists of several components: a retinal component that scales its output based on the mean and variance of the input (similarly to RGCs in the retina); a membrane potential component which also tracks the mean membrane potential; and a threshold component which works together with the membrane potential component to determine the neuron activation at each step. All of these components are tied together by a single parameter (denoted as $ \tau $), which is analogous to the membrane time constant. Note that although some observations are made about the biological equivalence of the model parameter and the membrane time constant, this parameter does not necessarily have a strict biological interpretation.

\subsection{Retinal model}\label{mpath.retina}
We begin with the observation that the retina adapts to both the mean and variance of the light intensity in the visual scene \parencite{Demb--2008--FunctionalCircuitryVisual,FreemanGranaPassaglia--2010--RetinalGanglionCell,MeisterBerry--1999--NeuralCodeRetina,BarlowLevick--1969--ChangesMaintainedDischarge,BaccusMeister--2002--FastSlowContrast,KimRieke--2001--TemporalContrastAdaptation,RiekeRudd--2009--ChallengesNaturalImages,RiderHenningStockman--2019--LightAdaptationControls}. This adaptation has two main purposes, namely contrast enhancement and dynamic range adaptation. Contrast enhancement involves BCs adapting their center/surround response characteristics to encode over the spatial extent of the receptive field, whereas dynamic range adaptation involves tracking the mean light intensity over time to ensure that the retina remains responsive as the mean illumination varies over nine orders of magnitude from midnight to midday and back.

The mean $ V_{\mu} $ and variance $ V_{\sigma^{2}} $ of a time-dependent variable $ V $ can be tracked in a step-wise manner with an exponential running average defined by a single decay parameter $ \alpha $. Formally, the mean and variance at step $ t_{k} $ are defined recursively as

\begin{align}
	V_{\mu}(t_{k}) & = V_{\mu}(t_{k-1}) + \alpha(V(t_{k}) - V_{\mu}(t_{k-1})) \label{eq.mean} \\
	V_{\sigma^{2}}(t_{k}) & = (1 - \alpha)(V_{\sigma^{2}}(t_{k}) + \alpha(V(t_{k}) - V_{\mu}(t_{k-1}))^{2}) \label{eq.var}
\end{align}

Here, $ V_{\mu}(0) = V_{\sigma^{2}}(0) = 0 $ and $ \alpha \in (0, 1) $. In the case of the retina, the variable being tracked is background luminance. Clearly, the choice of $ \alpha $ affects the speed of adaptation, where a larger value results in faster adaptation and vice versa.

In the MPATH model, the mean and variance are used to obtain a normalised estimate $ \Delta \tilde{V}(t_{k}) $ of the deviation from the mean value of the variable at time step $ t_{k} $:
\begin{align}
	\Delta \tilde{V}(t_{k}) = \frac{V(t_{k}) - V_{\mu}(t_{k - 1})}{\sqrt{V_{\sigma^{2}}(t_{k - 1})}}
\end{align}

Next, the graded normalised response $ \rho_{ON/OFF} $ of idealised ON- and OFF-type RGCs at step $ t_{k} $ is obtained by passing $ \Delta \tilde{V}(t_{k}) $ through two rectified sigmoid ($ \tanh $) transfer functions mirrored about the ordinate, which model the respective responses of ON and OFF-type RGCs in the retina:

\begin{align}
	\rho_{ON}(t_{k}) & =
	\begin{cases}
		\tanh( \Delta \tilde{V}(t_{k}) ) & \Delta \tilde{V}(t_{k}) > 0 \\
		0 & otherwise
	\end{cases} \\
	\rho_{OFF}(t_{k}) & =
	\begin{cases}
	0 & \Delta \tilde{V}(t_{k}) > 0 \\
	\tanh( \lvert\Delta \tilde{V}(t_{k})\rvert ) & otherwise
	\end{cases} \label{eq.on_off.response}
\end{align}

Note that we use the absolute value of $ \Delta \tilde{V} $ in the case of OFF RGCs in order to obtain a non-negative response (the spiking rate cannot be negative). Using such interleaved rectified sigmoid activation is justified from a theoretical perspective \parencite{Laughlin--1981--SimpleCodingProcedure} and is also supported by observations of the response of ON- and OFF- RGCs \parencite{BarlowFoldiak--1989--AdaptationDecorrelationCortex,RatliffBorghuisKaoEtAl--2010--RetinaStructuredProcess}. This model is similar to the one in \parencite{Hadjiivanov--2016--AdaptiveConversionRealvalued}, but with two key differences. First, the response is based on mirrored rectification rather than the difference of the responses of ON and OFF-type RGCs. This makes sense from a biological perspective because although ON- and OFF-type RGCs are often paired in the retina \parencite{Schiller--2010--ParallelInformationProcessing}, the ON and OFF pathways are kept separate until they converge in the visual cortex \parencite{ZaghloulBoahenDemb--2003--DifferentCircuitsRetinal}. The second difference, which is a consequence of the first one, is that this model automatically doubles the number of neurons in the input (retinal) layer as there are now two separate (ON and OFF) channels for each input. Although this model ignores the low but consistent maintained spike discharge present in the retina even in complete darkness \parencite{BarlowFitzhughKuffler--1957--DarkAdaptationAbsolute}, this `minimal spiking rate' can be taken into account, for example, by scaling the activation using a term $ \gamma < 1$ and adding $1 - \gamma $ to each response $ \rho_{ON/OFF} $. In either case, the merit of this approach is that the response of the retina is capped at $ 1 $, which represents a normalised maximal spiking rate $ s_{max} $ that can be set to an arbitrary value.

There still remains the question of the value of $ \alpha $ in \cref{eq.mean,eq.var}. In biological neurons, the membrane potential rises and decays at a rate proportional to the inverse of the membrane time constant:

\begin{align}
  \tau_{m} ( V_{rise}(t) & = V_{0}(1 - e^{-\frac{t}{\tau_{m}}}) \\
  V_{decay}(t) & = V_{0}e^{-\frac{t}{\tau_{m}}}
\end{align}

where $ V_{0} $ is the initial potential), without loss of generality we set $ \alpha = \frac{1}{\tau_{m}} $.

The response of the retina is graded (sustained) and lacks a threshold. Therefore, the retinal response is paired with an adaptive activation threshold model for downstream neurons in the visual cortex.

\subsection{Adaptive activation threshold}\label{mpath.threshold}
As outlined in Section \ref{mpath.background}, the spiking threshold $ \theta $ follows relatively slow fluctuations in the mean membrane potential while responding with a sudden \textit{drop} to strong membrane depolarisation. Evidence from both \textit{in vivo} experiments and simulations demonstrates a strong positive correlation between the spiking threshold $ \theta $ and the mean membrane potential $ V_{\mu} $ as well as an inverse correlation between the \textit{speed} of depolarisation immediately preceding a spike. Specifically, the time evolution of the spiking threshold can be modelled similarly to that of the membrane potential itself \parencite{FontainePenaBrette--2014--SpikeThresholdAdaptationPredicted,AzouzGray--2003--AdaptiveCoincidenceDetection,PlatkiewiczBrette--2010--ThresholdEquationAction,PlatkiewiczBrette--2011--ImpactFastSodium}:

\begin{equation}
	\tau_{\theta}\frac{d\theta}{dt} = \theta_{\infty}(V_{m}) - \theta \label{eq.fontaine.theta}
\end{equation}

Here, $ V_{m} $ is the membrane potential, $ \theta_{\infty}(V_{m}) $ is the steady-state threshold and $ \tau_{\theta} $ is the time constant of the threshold adaptation.

In contrast to the well-established limits of the above parameters in biological neurons, in an idealised model the threshold must adapt to a varying membrane potential with \textit{a priori} unknown parameters such as the steady-state threshold $ \theta_{\infty}(V_{m}) $ and $ \tau_{\theta} $. Therefore, we adopt the same normalisation and scaling technique applied to RGCs above in order to obtain a predictable output range. Specifically, the membrane potential $ \rho(t_{k}) $ at time $ t_{k} $ is taken as $ \rho(t_{k}) = \tanh(\Delta \tilde{V}(t_{k})) $. Note that there is no rectification here as the neuron's membrane potential is allowed to dip below the mean potential. Importantly, the fact that $ \rho $ is capped at $ 1 $ allows us to set the steady-state threshold $ \theta_{\infty}(V_{m}) $ to $ 1 $ for times \textit{up to the current time step}, which still guarantees the threshold to be \textit{above} the membrane potential.

Next, we use the following relation \parencite{AzouzGray--2000--DynamicSpikeThreshold} between the change in threshold and the rate of membrane depolarisation over a brief integration period (a single time step):

\begin{equation}
	\theta = a + be^{-\frac{\frac{dV_{m}}{dt}}{c}} \label{eq.aozuz.theta}
\end{equation}

Here, $ V_{m} $ denotes the membrane potential. Based on the above assumptions that the threshold is set to track the mean membrane potential, the input is normalised, the steady-state threshold is fixed at $ 1 $ and the time constant of threshold adaptation is $ \tau_{\theta} $, in \cref{eq.aozuz.theta} we set $ a = 0 $, $ b = 1 $, and $ c = \tau_{\theta} $. Finally, with the assumption that the input is normalised (using \cref{eq.mean,eq.var}), the rate of increase $ \dot V_{m} $ of them membrane potential over a unit time step $ \Delta t $ (where we take $ \Delta t = t_{k} - t_{k - 1} = 1 $) becomes simply

\begin{equation}
	\dot V_{m}(t_{k}) = \frac{\Delta \tilde{V}(t_{k})}{\Delta t} = \frac{V_{m}(t_{k}) - V_{\mu}(t_{k - 1})}{V_{\sigma}(t_{k - 1})},
\end{equation}

where $ V_{\sigma} = \sqrt(V_{\sigma^{2}})$ is the standard deviation of the input. Formally, the spike threshold at time $ t_{k} $ is then described as follows:

\begin{align}
	\theta_{\infty}(V_{m}) & = \theta_{t < t_{k}} = 1 \\
	\theta(t_{k}) & = e^{ - \frac{\Delta \tilde{V}(t_{k})}{\tau_{\theta}}}
\end{align}

Finally, we can compute the thresholded activation $ \rho_{\theta} $ of the neuron as the difference of the membrane potential and the threshold:

\begin{equation}
	\rho_{\theta}(t_{k}) = \rho(t_{k}) - \theta(t_{k}) = \tanh(\Delta \tilde{V}(t_{k})) - e^{ - \frac{\Delta \tilde{V}(t_{k})}{\tau_{\theta}}}
\end{equation}

\begin{figure}
	\centering
	\begin{tikzpicture}
		\begin{axis}%
		[%
		yscale=0.5,
		xscale=1.5,
		axis x line=bottom,
		axis y line=center,
		xtick={0},
		xticklabels={},
		ytick={1},
		yticklabels={},
		ymax=1.25,
		xmin=-3,
		xmax=3,
		xlabel=,
		xlabel style={at={(current axis.right of origin)},anchor=north},
		legend style={at={(0.7,0.7)},anchor=east,fill=none,draw=none},
		yticklabel style={anchor=south east},
		domain=-3:3
		]
		\addplot%
		[%
		blue,
		mark=none,
		samples=500,
		smooth,
		thick,
		domain=0:3
		]
		(x,{tanh(x)});
		\addplot%
		[%
		black!50!green,
		mark=none,
		samples=100,
		smooth,
		domain=0:3,
		]
		(x,{exp(-x)});
		\legend{\tiny{$ tahn\left(\Delta \tilde{V}(t_{i})\right) $},
		\tiny{$ exp\left(-\frac{\Delta \tilde{V}(t_{i})}{\tau_{\theta}}\right) $}}
		\addplot
		[%
		black,
		mark=none,
		samples=2,
		domain=-3:3,
		thick,
		dotted
		]
		{1};
		\end{axis}
		\node at (4.9,2.5) {\tiny{1}};
		\node at (5.2,-0.3) {\tiny{$\mu_{i}$}};
	\end{tikzpicture}
\end{figure}

The neuron fires only when this difference is positive, in other words, when the membrane potential has reached the threshold. Since we are assuming that the neuron acts as a temporal integrator, we define $ \eta $ as the activation rate of the neuron relative to its saturation level (maximum activation rate):

\begin{equation}\label{eq.eta}
	\eta(t_{k}) =
	\begin{cases}
		\rho_{\theta}(t_{k}) & \rho_{\theta}(t_{k}) > 0 \\
		0 & otherwise
	\end{cases}
\end{equation}

Apart from the presence of a threshold, another difference between downstream neurons and the retina in this model is that the membrane potential $ V_{m} $ is reset to $ 0 $ if $ \eta(t_{k}) > 0$ in order to mirror the behaviour of biological neurons.

As the activation rate of neurons tends to decay exponentially over time \parencite{TiganjHasselmoHoward--2015--SimpleBiophysicallyPlausible}, we can go one step further and define an exponential decay for $ \eta $, in which case \cref{eq.eta} should be rewritten as:

\begin{align}
	\eta(t_{0}) & = 0 \\
	\eta(t_{k}) & =
	\begin{cases}
		\rho_{\theta}(t_{k}) & \rho_{\theta}(t_{k}) > 0 \\
		\eta(t_{k - 1})e^{-\frac{1}{\tau_{\eta}}} & otherwise
	\end{cases}
\end{align}

where $ \tau_{\eta} $ is the time constant of the activation rate decay.

Clearly, setting the steady-state threshold to a (non-negative) value other than $ 1 $ will not have an impact on the fact that the activation rate as defined above is confined between $ 0 $ and $ 1 $. Intuitively, although a different value for the steady-state threshold \textit{would} affect the intercept between the rise of the membrane potential $ \rho $ and the threshold $ \theta $, this can be compensated for by proper scaling of the membrane time constant. A proper investigation of this is left for future research.

In this regard, it has been determined that a plausible value for the threshold time constant $ \tau_{\theta} $ is an order of magnitude \textit{smaller} than the membrane time constant $ \tau_{m} $ \parencite{FontainePenaBrette--2014--SpikeThresholdAdaptationPredicted}. This makes sense from the point of view that it allows the threshold to adapt much faster than the membrane itself in order to pick up (and propagate) deviations from the mean faster than the membrane itself. This also reinforces the above assumption of a dynamic change in threshold \textit{within} as single time step. Therefore, the default value of $ \tau_{\theta} $ is set to $ \frac{\tau_{m}}{10} $, and $ \tau_{\eta} $ is set to match $ \tau_{\theta} $.

Functionally, the threshold acts as a `variance sieve' that filters out signals that are close to the mean potential while letting through the outliers which deviate strongly from the mean. In other words, the threshold acts as a decision boundary for the question \textit{Is this news?} \parencite{Barlow--1961--PossiblePrinciplesUnderlying}, whereby redundant information is filtered out by progressively tougher `critics' as it flows through a series of layers of thresholded neurons. This is also consistent with the observation that the neuron homeostatically maintains a constant difference between the average membrane potential and the threshold, whereby ``[...] the firing rate becomes less and less sensitive to the input mean [...] and relatively more sensitive to the variance'' \parencite{PlatkiewiczBrette--2011--ImpactFastSodium}.

\subsection{Synaptic scaling}\label{mpath.synscale}
In biological neurons, synaptic scaling acts as a homeostatic mechanism that allows LTP and LTD to proceed without causing uncontrolled changes in the firing rate of the neuron \parencite{Turrigiano--2008--SelfTuningNeuronSynaptic,HartmannMinerTriesch--2016--PreciseSynapticEfficacy}. As a rule based on positive feedback, Hebbian learning can be unstable, leading to silencing or runaway excitation \parencite{ZenkeHennequinGerstner--2013--SynapticPlasticityNeural,AbbottNelson--2000--SynapticPlasticityTaming}. Therefore, although synaptic scaling itself is (at least computationally) a fairly straightforward procedure, it is important as a control mechanism applied in tandem with Hebbian learning to ensure that weights remain within healthy limits.

The MPATH model implements synaptic scaling together with the generalised Hebbian learning (GHL) rule, which is applicable to multi-neuron layers. GHL takes the following form \parencite{Sanger--1989--OptimalUnsupervisedLearning}:

\begin{equation}\label{eq.ghl}
	\Delta\boldsymbol{W}(t) = \lambda \left( \boldsymbol{y}(t)\boldsymbol{x}^{T}(t) - \boldsymbol{LT}\left[ \boldsymbol{y}(t)\boldsymbol{y}^{T}(t) \right]\boldsymbol{W}(t) \right),
\end{equation}

where $ \boldsymbol{W} $ is the weight matrix between a layer of input neurons $ \boldsymbol{x} $ and a layer of output neurons $ \boldsymbol{y} $, $ \lambda $ is a small learning rate (where $\lim_{t \to \infty}(\lambda(t)) = 0$ and $ \sum_{0}^{\infty}(\lambda(t)) = \infty $) and $ \boldsymbol{LT}[\boldsymbol{M}] $ represents the lower triangular submatrix of matrix $ \boldsymbol{M} $. At each step where GHL is applied, synaptic scaling normalises the input weight vector for each neuron:

\begin{equation}\label{eq.synscale}
	\boldsymbol{W_{j}}' = \frac{\boldsymbol{W_{j}}}{\norm{\boldsymbol{W_{j}}}},
\end{equation}

where $ \norm{\boldsymbol{W_{j}}} $ is the Euclidean norm of row $ j $ in $ \boldsymbol{W} $.

\section{Experimental verification}\label{mpath.experiments}

The MPATH model was implemented in TensorFlow using custom layers and tensor operations. Computing neuron activations is efficient as the only parameter that the model depends on is the membrane time constant $ \tau_{m} $ and the running mean and variance can be tracked with minimal overhead.

The response of the model with a three-layer network was tested using several types of time-varying input in order to establish whether the network can adapt to and learn from the input as expected. The parameters of the network were the same in all three experiments (Table \ref{tbl.params}), with the exception of the pattern being presented and the status of Hebbian learning, which are indicated below.

\begin{table}
	\caption{Parameters for the experiments presented in Figs. \ref{fig.noise.no_learning} -- \ref{fig.sine_wave_grating}.}
	\label{tbl.params}
	\centering
	\footnotesize
	\begin{tabularx}{\textwidth}{X|X}
		\toprule
		Parameter                                                                                  & Value                                                    \\
		\midrule
		Input size                                                                                 & $ 30 $ channels                                          \\
		Retinal layer size                                                                         & $ 60 $ ($ 30 $ ON and $ 30 $ OFF)                        \\
		Layer 1 size                                                                               & $ 20 $                                                   \\
		Layer 2 size                                                                               & $ 10 $                                                   \\
		Weights\xdef\tempwidth{\the\linewidth}                                                    & Uniformly initialised $ \in (-1,1) $                     \\
		\multicolumn{1}{m{\tempwidth}|}{Membrane time constant $ \tau_{m} $ \newline (retinal layer)}        & $ 100 $                                                  \\
		\multicolumn{1}{m{\tempwidth}|}{Membrane time constant $ \tau_{m} $ \newline (layers 1 and 2)}       & \multicolumn{1}{m{\tempwidth}}{$ 5 $ -- $ 25 $ \newline (distributed evenly across all neurons)} \\
		\multicolumn{1}{m{\tempwidth}|}{Threshold time constant $ \tau_{\theta} $ \newline (layers 1 and 2)} & $ 0.1 \tau_{m} $                                       \\
		Activation decay time constant $ \tau_{\eta} $ (layers 1 and 2)                            & $ \tau_{\theta} $                                      \\
		Duration of simulation                                                                     & $ 2000 $ steps                                           \\
		Learning rate (Hebbian learning)                                                           & $ 0.05 $ (multiplied by $ 0.99 $ every 10 steps)         \\
		\bottomrule
	\end{tabularx}
\end{table}

\begin{figure}[htbp!]
	\begin{subfigure}{0.95\textwidth}
		\centering
		\includegraphics[width=\textwidth]{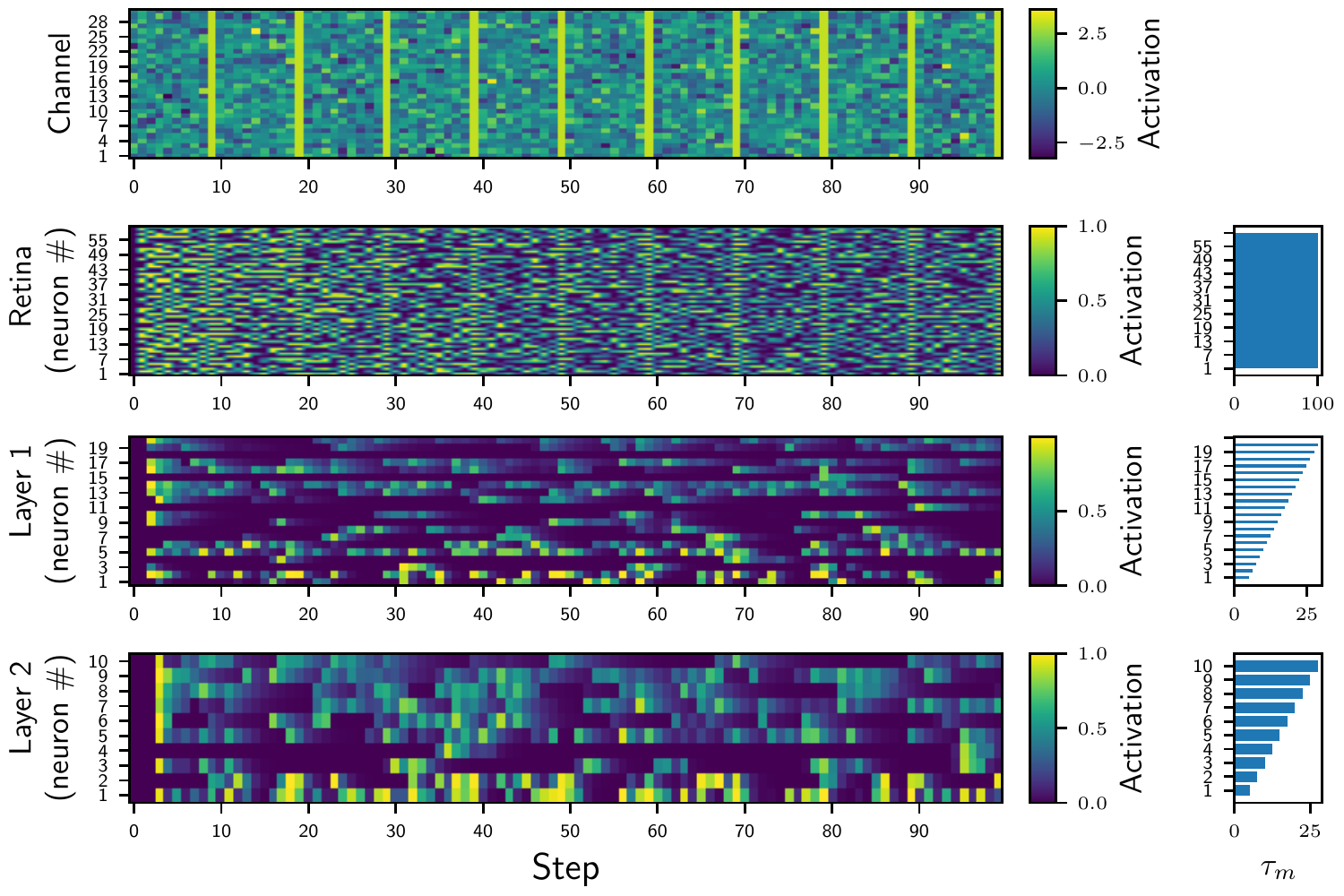}
		\subcaption{\label{fig.noise.no_learning.0-100}}
	\end{subfigure}
	\hfill
	\begin{subfigure}{0.95\textwidth}
		\centering
		\includegraphics[width=\textwidth]{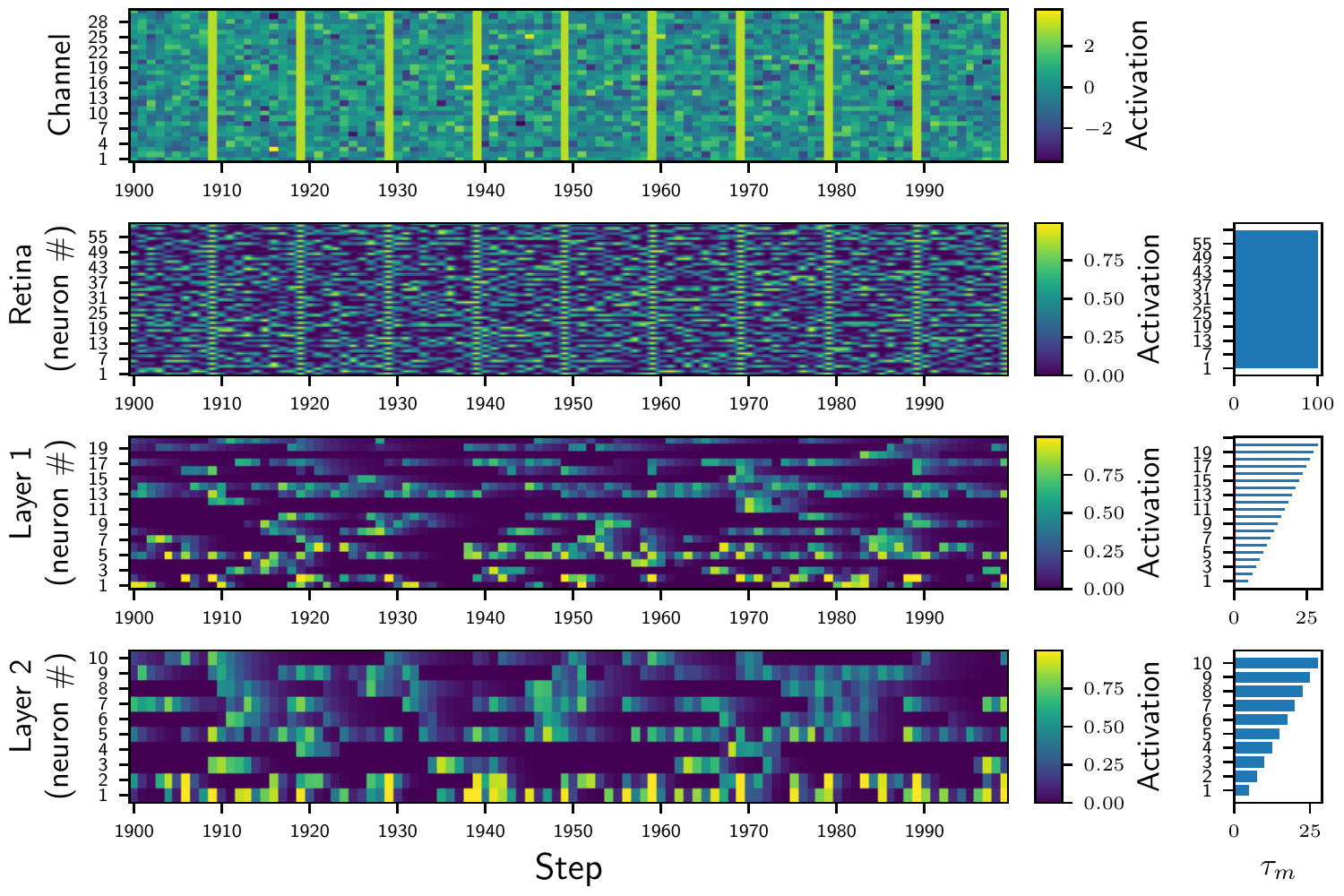}
		\subcaption{\label{fig.noise.no_learning.1900-2000}}
	\end{subfigure}
	\caption{\label{fig.noise.no_learning}
		Raster plots of the activations of a three-layer network presented with Gaussian noise for $ 2000 $ steps, with flashes of random magnitude superimposed every $ 10 $ steps. (\subref{fig.noise.no_learning.0-100}) steps $ 0 $--$ 100 $ and (\subref{fig.noise.no_learning.1900-2000}) steps $ 1900 $--$ 2000 $. Hebbian learning is disabled. Clearly, although the network does not learn, it can still adapt to the input, with decaying memory traces of the activation seen after each action potential.}
\end{figure}

\begin{figure}[htbp!]
	\begin{subfigure}{0.95\textwidth}
		\centering
		\includegraphics[width=\textwidth]{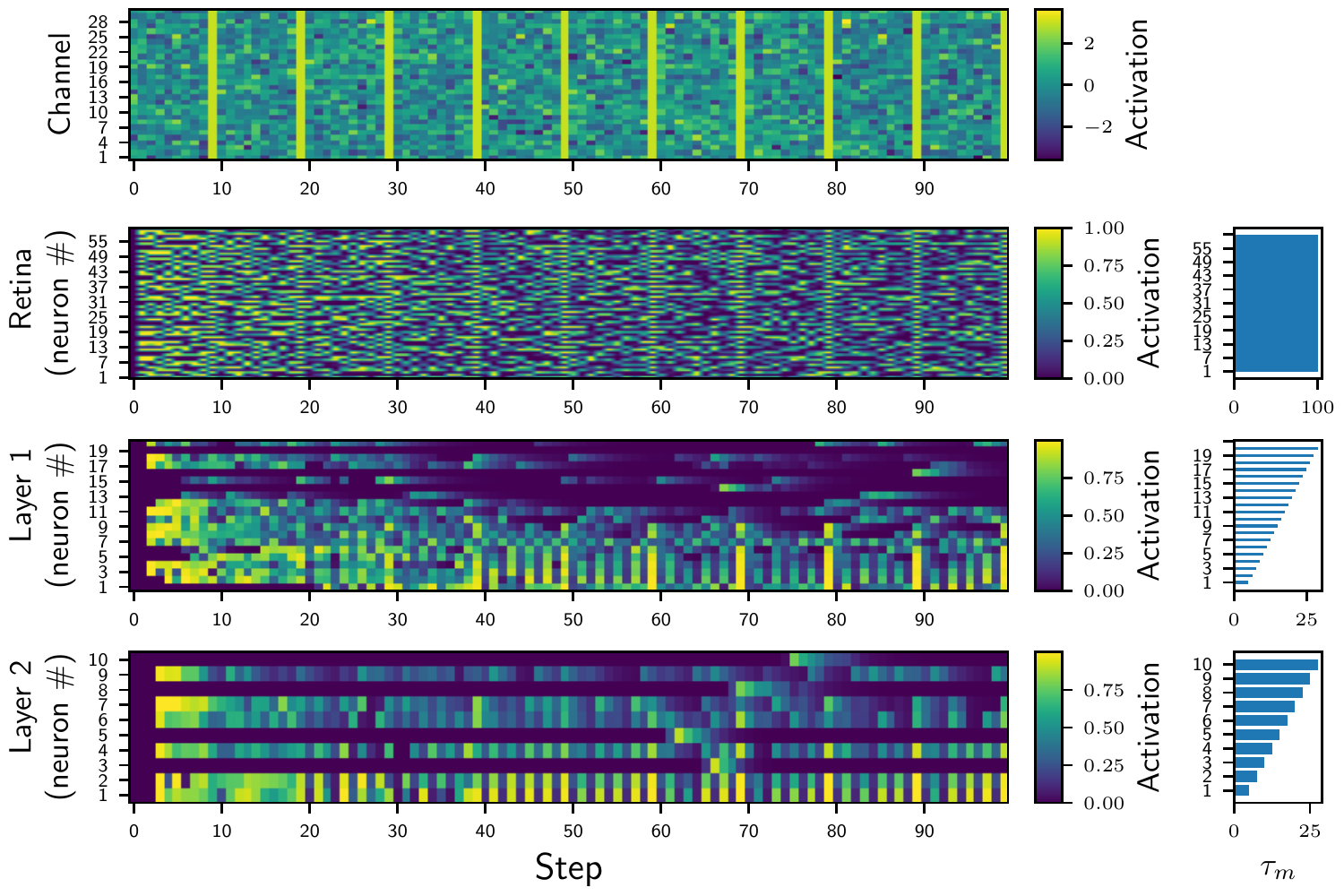}
		\subcaption{\label{fig.noise.learning.0-100}}
	\end{subfigure}
	\hfill
	\begin{subfigure}{0.95\textwidth}
		\centering
		\includegraphics[width=\textwidth]{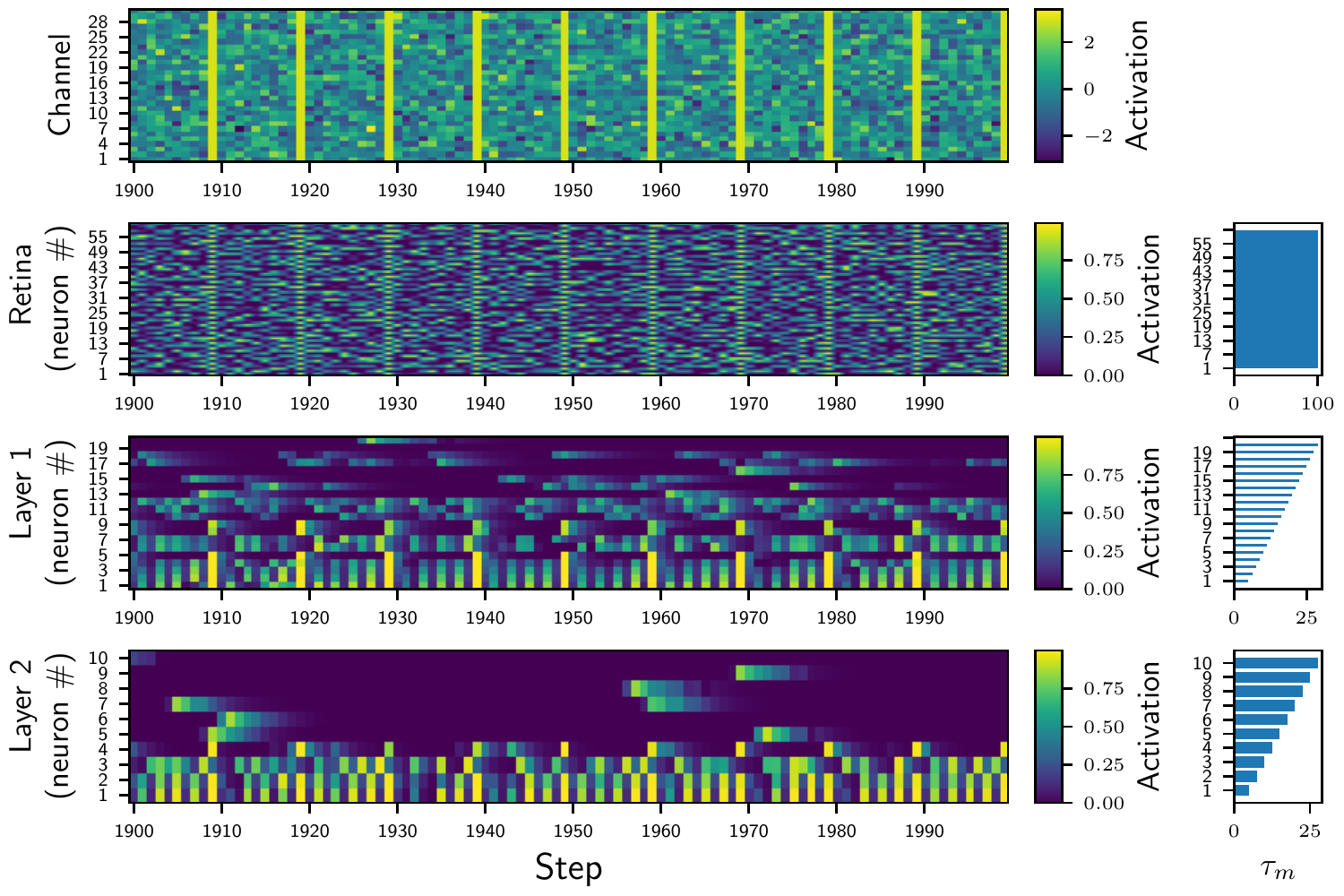}
		\subcaption{\label{fig.noise.learning.1900-2000}}
	\end{subfigure}
	\caption{\label{fig.noise.learning}
		Plots for the same conditions as in Fig. \ref{fig.noise.no_learning}, but with Hebbian learning enabled. The network not only adapts to the input, but also learns from it, with the two processes complementing each other. Most of the uninformative input (noise) is filtered out, while the important information (the flashes) are clearly represented in both layers. Memory traces of action potentials in layers $ 1 $ and $ 2 $ are clearly visible in this case as well.}
\end{figure}

The figures show raster plots similar to those used for visualising the response of each network layer. In Figs. \ref{fig.noise.no_learning} and \ref{fig.noise.learning}, the network is presented with Gaussian noise with a mean of $ 0 $ and an SD of $ 1 $, with intermittent flashes with a magnitude of $ 3 $ superimposed onto the noise every $ 10 $ steps. In Fig. \ref{fig.sine_wave_grating}, the network is trained on a sharp sine-wave grating pattern applied with a spatial period of $ 1 $ channel and a temporal period of $ 2 $ steps (i.e., the grating is shifted one channel up every two steps, so that illuminated areas become dark and vice versa; the grating wraps around). Note that due to the pairing of ON and OFF RGCs in the retinal layer the number of inputs is double that of the input channels. The bar charts next to each layer show the membrane time constants $ \tau_{m} $ for the neurons in that layer.

Figure \ref{fig.noise.no_learning} shows the results when Hebbian learning was disabled. While the network is responsive and demonstrates the ability to adapt to the input, it does not produce any noticeable activation pattern in response to the superimposed flashes. The situation is very different in Fig. \ref{fig.noise.learning}, where the presented pattern is the same but with Hebbian learning enabled. In this case, the network quickly learns to filter out the noise and produces a much more regular activation pattern in response to the superimposed flashes (the patterns in layers $ 1 $ and $ 2 $ are already noticeable by step $ 100 $).

In Fig. \ref{fig.sine_wave_grating.0-100}, the network is presented with Gaussian noise with a mean of $ 0 $ and an SD of $ 1 $ for the first $ 50 $ steps, followed by the continuous presentation of the grating pattern (magnitude of dark and light sections of $ 0 $ and $ 1 $, respectively) until step $ 1000 $. The network responds with chaotic firing at first, but quickly starts to learn the spatiotemporal pattern. By step $ 1000 $ (Fig. \ref{fig.sine_wave_grating.950-1050}), the network is firing regularly in response to the pattern. At step $ 1000 $, the grating pattern is switched back to Gaussian noise, and training is stopped. At step $ 1500 $ (Fig. \ref{fig.sine_wave_grating.1450-1550}), the grating pattern is switched back on, but training is \textit{not} resumed. The network quickly recovers from the random input and resumes a firing pattern very close to the one learned in the first $ 1000 $ steps, without any additional training (Fig. \ref{fig.sine_wave_grating.1900-2000}). This demonstrates that learned behaviour is not forgotten in the process of adaptation to a different input pattern. It should be noted that there are no recurrent connections in any of the trials. The network retains a memory of what it has experienced in the recent past purely by using the internal states of its neurons.

It is noteworthy that in all cases the network has a `cold start', meaning that it has no prior knowledge of the range of values that the input can take. Instead, each neuron adapts dynamically to changes in the input while retaining its sensitivity to deviations from the mean and conveys those deviations only when the threshold is crossed. Also, the fading traces of activation after each action potential show that neurons retain a short-lived internal memory of the input, even in the absence of recurrent connections.

\section{Conclusion}\label{mpath.discussion}
The MPATH model demonstrates the flexibility and effectiveness of homeostatic processes found in biological neurons. Importantly, it explores the possibility of harnessing the underutilised internal state of neurons in a neural network as a way to adapt to the input while learning from it. As seen in the experiments above, adaptation and unsupervised learning go hand in hand, which posits some exciting questions about the nature of learning and adaptation in biological brains. One such question is whether parameters such as the membrane time constant, the activation threshold and the activation decay time constant have to be strictly within certain limits that are tightly controlled by the neuron, or whether they are artefacts of evolution and adaptation in particular environments and for particular functions. Another question is whether these time constants can be derived from first principles such as variance in the input data, or whether they themselves can be \textit{learned} in an unsupervised manner.

This model is part of broader framework for adaptive unsupervised deep learning which is under active development. There are many simplifications in this proof of concept that need to be addressed in future research. The most obvious one is the lack of a \textit{spatial} adaptation component in the retinal layer. In this regard, some intricate details about the organisation, size and function of RGCs in the retina must be considered. For instance, it has been shown that the retina is more sensitive to negative contrast (dark spot on light background) than to positive contrast (light spot on dark background), a fact mirroring the imbalance between negative and positive contrast in natural scenes\parencite{MazadeJinPonsEtAl--2019--FunctionalSpecializationCortical,RatliffBorghuisKaoEtAl--2010--RetinaStructuredProcess}. In addition, even though OFF cells are physically smaller than ON cells and have a smaller a receptive field\parencite{ChichilniskyKalmar--2002--FunctionalAsymmetriesGanglion}, both ON and OFF cells cover the entire visual field, meaning that OFF cells outnumber ON cells by a factor of about $ 1.7 $ \parencite{RatliffBorghuisKaoEtAl--2010--RetinaStructuredProcess,Dacey--1993--MosaicMidgetGanglion}. A retinotopic convolutional layer that incorporates these and other properties of the retina is being developed to be able to explore both spatial as well as temporal adaptation within the MPATH model.

\newpage

\printbibliography
\newpage
\appendix
\section{Supplementary figures}\label{mpath.supplementary_figures}

\begin{figure}[htbp!]
	\begin{subfigure}{0.95\textwidth}
		\centering
		\includegraphics[width=\textwidth]{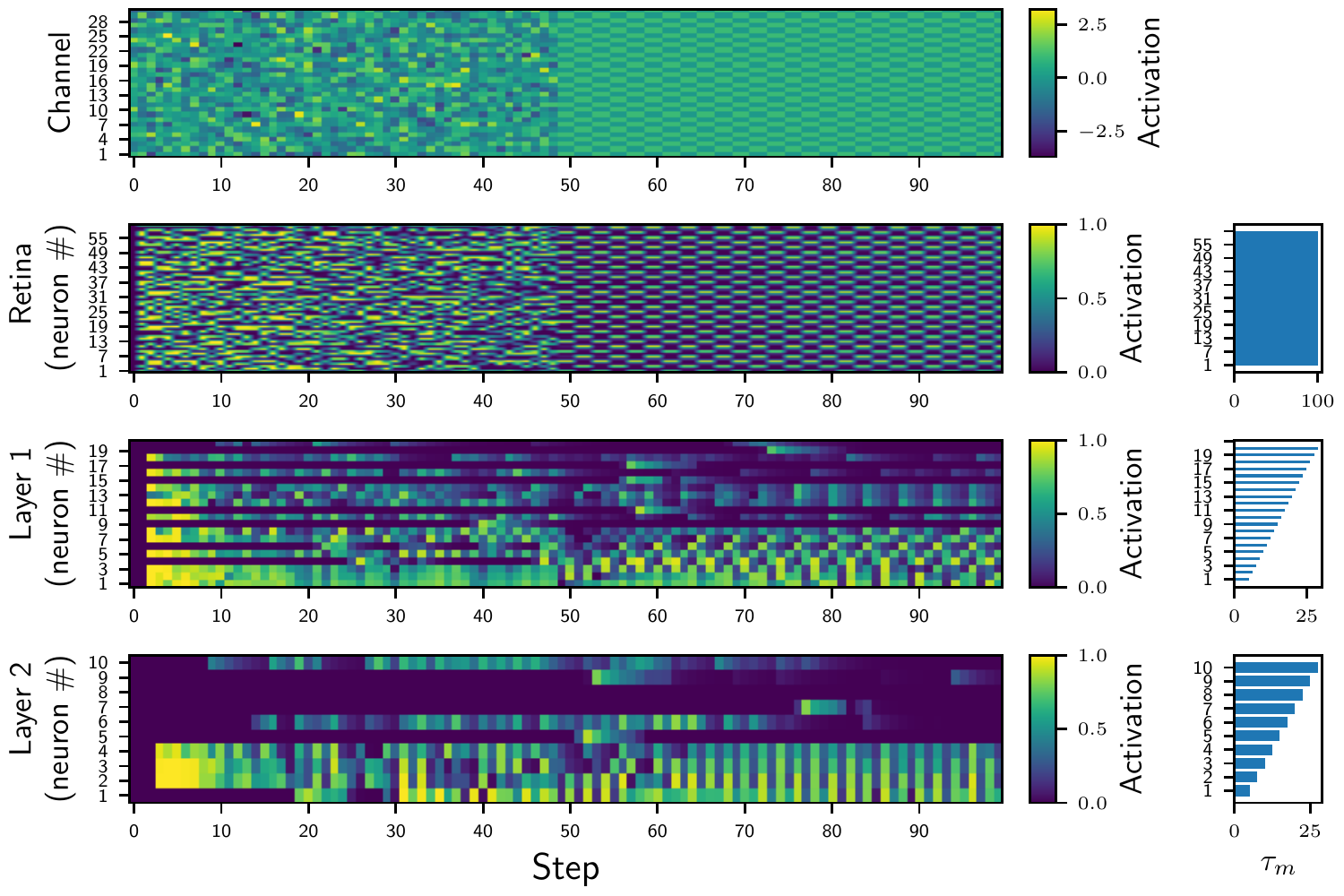}
		\subcaption{\label{fig.sine_wave_grating.0-100}}
	\end{subfigure}
	\hfill
	\begin{subfigure}{0.95\textwidth}
		\centering
		\includegraphics[width=\textwidth]{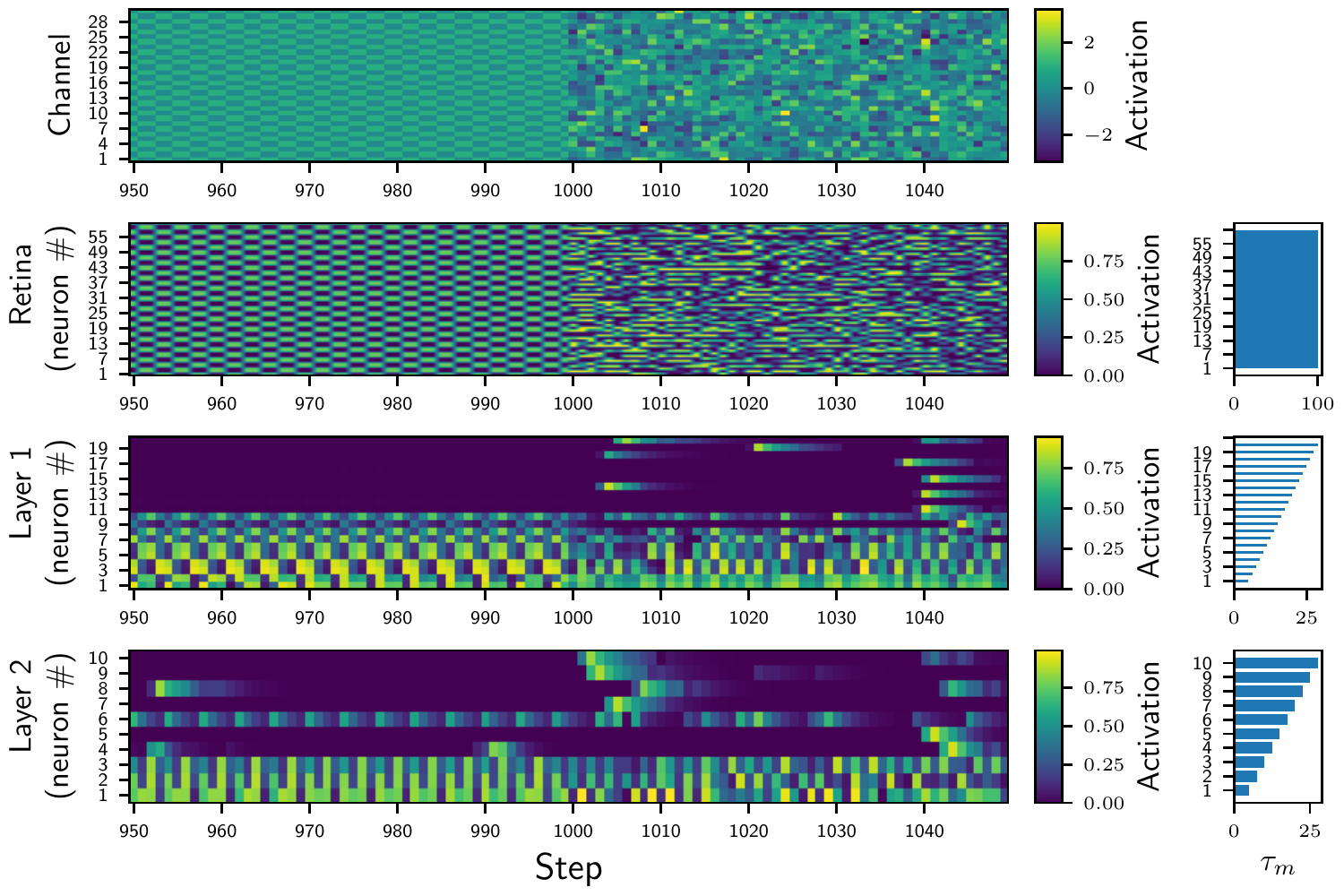}
		\subcaption{\label{fig.sine_wave_grating.950-1050}}
	\end{subfigure}
\end{figure}
\begin{figure}[htbp!]\ContinuedFloat
	\begin{subfigure}{0.95\textwidth}
		\centering
		\includegraphics[width=\textwidth]{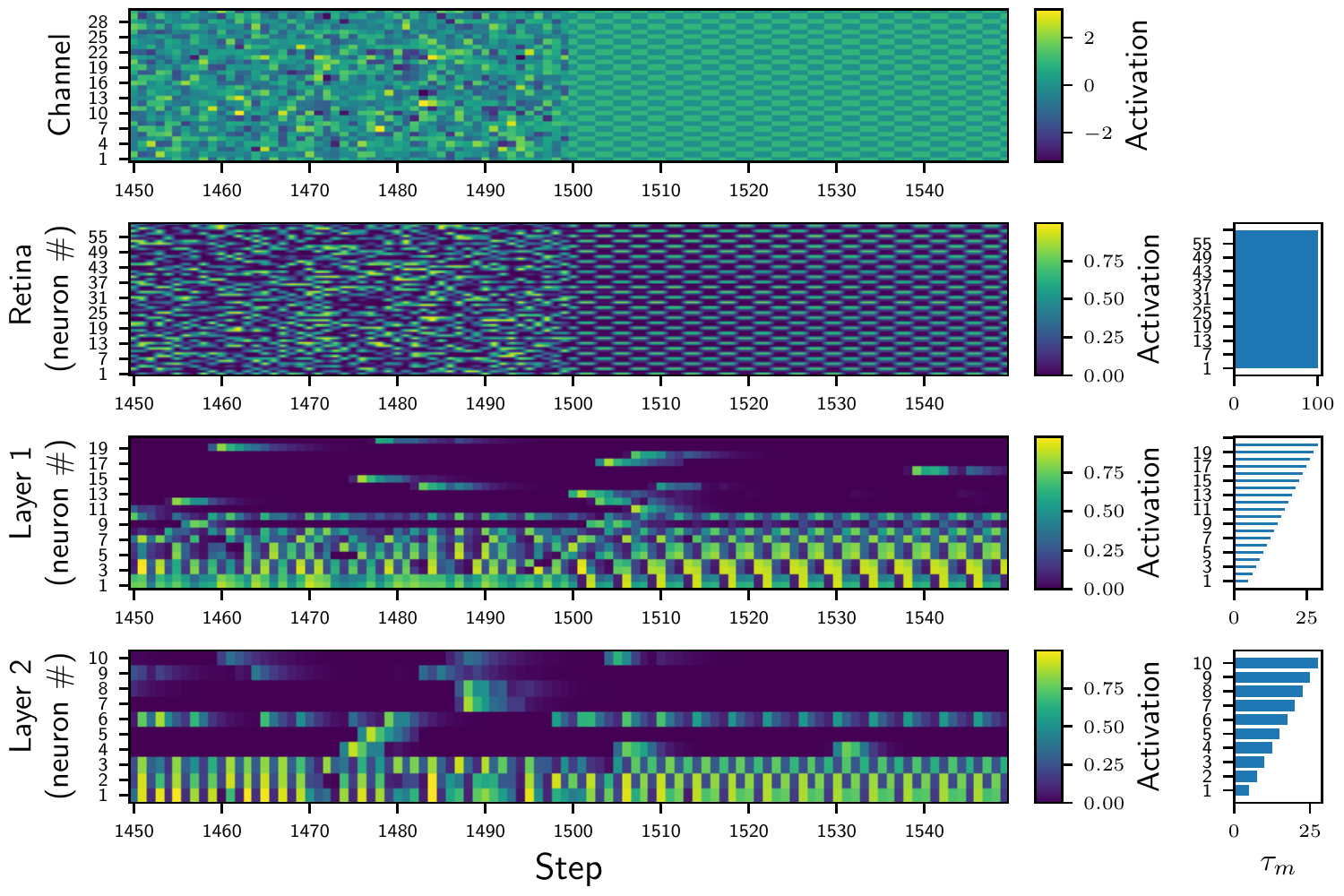}
		\subcaption{\label{fig.sine_wave_grating.1450-1550}}
	\end{subfigure}
	\hfill
	\begin{subfigure}{0.95\textwidth}
		\centering
		\includegraphics[width=\textwidth]{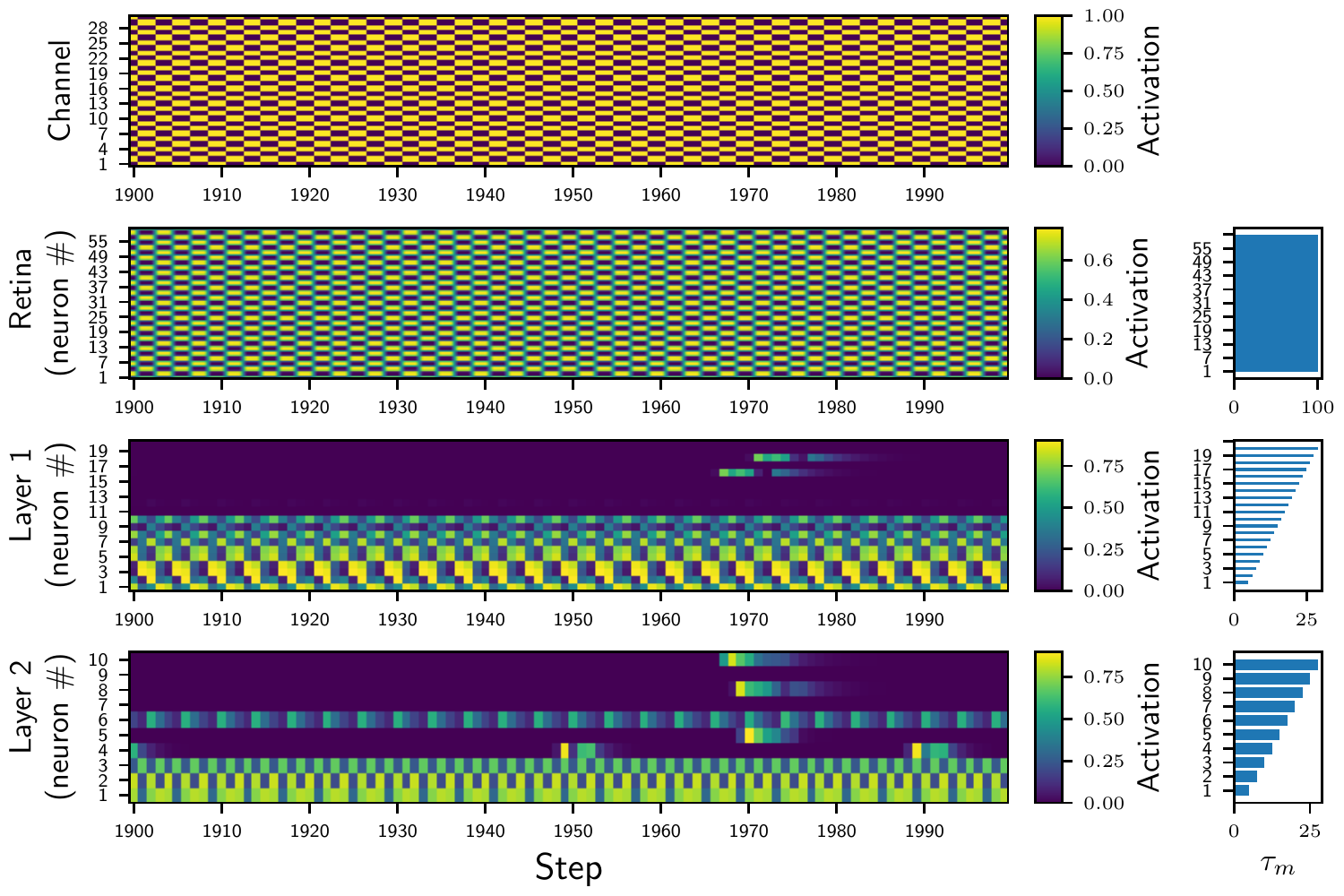}
		\subcaption{\label{fig.sine_wave_grating.1900-2000}}
	\end{subfigure}
	\caption{\label{fig.sine_wave_grating}
		Raster plots for a three-layer network trained on a sine-wave grating. (\subref{fig.sine_wave_grating.0-100}) The network is presented with random Gaussian noise for $ 50 $ steps, after which it is presented with a sine-wave grating. (\subref{fig.sine_wave_grating.950-1050}) The training is stopped half way through the trial (step $ 1000 $), and the network is presented with random noise for $ 500 $ steps. (\subref{fig.sine_wave_grating.1450-1550},\subref{fig.sine_wave_grating.1900-2000}) At step $ 1500 $, the sine-wave grating pattern is switched back on, but training remains disabled. The network recovers quickly and resumes a firing pattern very similar to that before the noise.}
\end{figure}

\begin{figure}[htbp!]
	\centering
	\begin{subfigure}{0.6\textwidth}
		\centering
		\includegraphics[width=\textwidth]{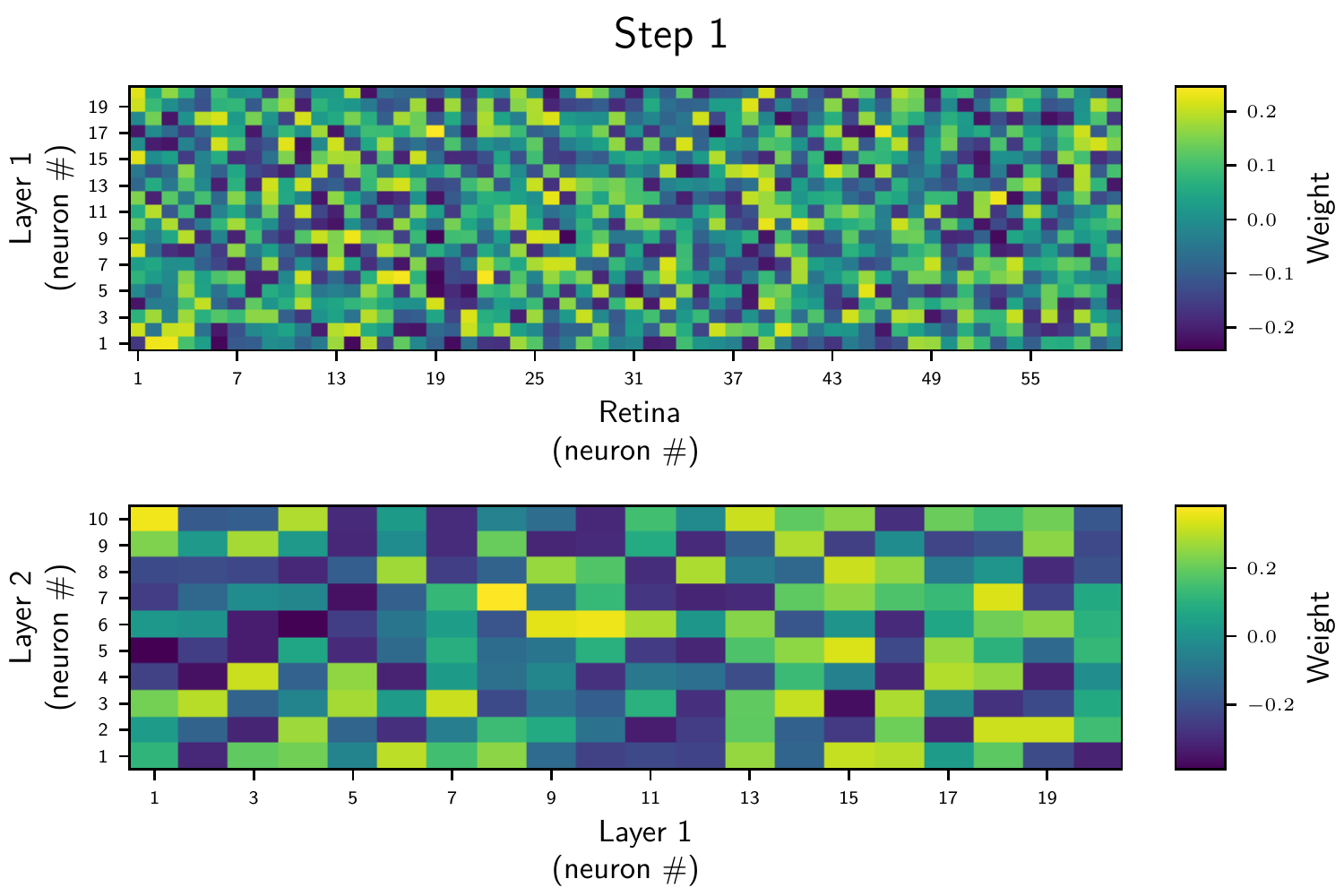}
		\subcaption{\label{fig.flash.weights.1}}
	\end{subfigure}
	\begin{subfigure}{0.6\textwidth}
		\centering
		\includegraphics[width=\textwidth]{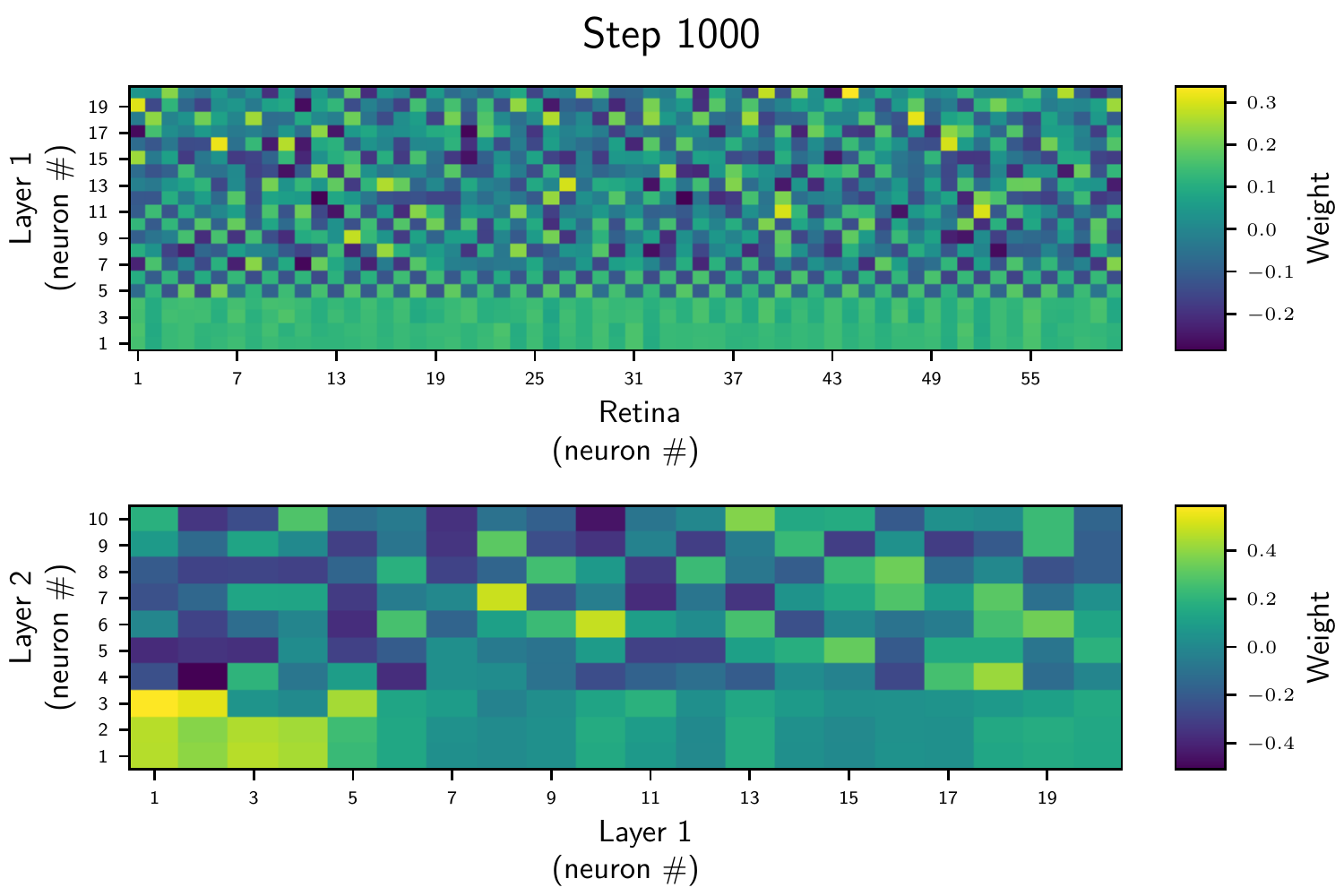}
		\subcaption{\label{fig.flash.weights.1000}}
	\end{subfigure}
	\begin{subfigure}{0.6\textwidth}
		\centering
		\includegraphics[width=\textwidth]{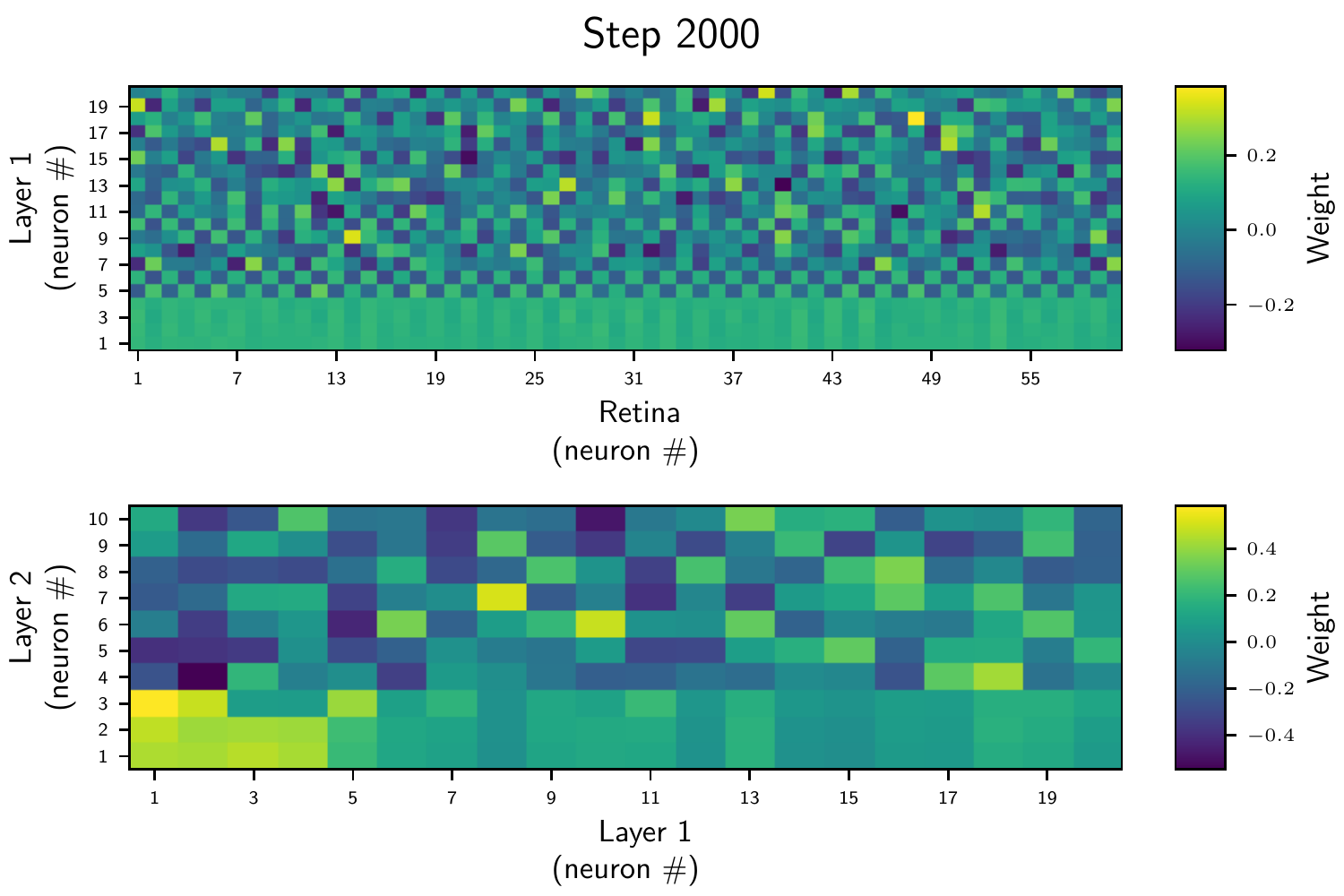}
		\subcaption{\label{fig.flash.weights.2000}}
	\end{subfigure}
	\caption{\label{fig.flash.weights}
		Weight matrix at steps $ 1 $, $ 1000 $ and $ 2000 $ of a three-layer network presented with Gaussian noise for $ 2000 $ steps, with flashes of random magnitude superimposed every $ 10 $ steps. (\subref{fig.noise.no_learning.0-100}) steps $ 0 $--$ 100 $ and (\subref{fig.noise.no_learning.1900-2000}) steps $ 1900 $--$ 2000 $. Hebbian learning is enabled. The weights have clearly stabilised by step $ 1000 $, with little noticeable difference between the weight matrix plots in steps $ 1000 $ and $ 2000 $.}
\end{figure}

\end{document}